\pdfoutput=1

\documentclass[11pt]{article}

\usepackage[]{acl}

\usepackage{times}
\usepackage{latexsym}
\usepackage[T1]{fontenc}
\usepackage[utf8]{inputenc}
\usepackage{microtype}
\usepackage{graphicx}
\usepackage{multirow}
\usepackage{booktabs}
\usepackage{threeparttable} 
\usepackage{tcolorbox}
\tcbuselibrary{breakable}
\usepackage{placeins}
\usepackage{bold-extra}
\usepackage{colortbl}
\definecolor{lightgray}{gray}{0.9}
\usepackage{float}
\usepackage{lscape}

\title{ELECTRA and GPT-4o:\\Cost-Effective Partners for Sentiment Analysis}

\author{James P. Beno \\
  Stanford Engineering CGOE \\
  \texttt{jim@jimbeno.net}
}

\begin{document}
\maketitle
\begin{abstract}

Bidirectional transformers excel at sentiment analysis, and Large Language Models (LLM) are effective zero-shot learners. Might they perform better as a team? This paper explores collaborative approaches between ELECTRA and GPT-4o for three-way sentiment classification. We fine-tuned (FT) four models (ELECTRA Base/Large, GPT-4o/4o-mini) using a mix of reviews from Stanford Sentiment Treebank (SST) and DynaSent. We provided input from ELECTRA to GPT as: predicted label, probabilities, and retrieved examples. Sharing ELECTRA Base FT predictions with GPT-4o-mini significantly improved performance over either model alone (82.50 macro F1 vs. 79.14 ELECTRA Base FT, 79.41 GPT-4o-mini) and yielded the lowest cost/performance ratio (\$0.12/F1 point). However, when GPT models were fine-tuned, including predictions decreased performance. GPT-4o FT-M was the top performer (86.99), with GPT-4o-mini FT close behind (86.70) at much less cost (\$0.38 vs. \$1.59/F1 point). Our results show that augmenting prompts with predictions from fine-tuned encoders is an efficient way to boost performance, and a fine-tuned GPT-4o-mini is nearly as good as GPT-4o FT at 76\% less cost. Both are affordable options for projects with limited resources.
\end{abstract}

\section{Introduction}

Sentiment analysis---the computational study of opinions, attitudes, and emotions in text \citep{medhat-2014-sentiment}---has seen major advances from transformer architectures \citep{vaswani-2017-attention}. Bidirectional encoders like BERT \citep{devlin-2019-bert}, RoBERTa \citep{liu-2019-roberta}, and ELECTRA \citep{clark-2020-electra} excel at sentiment analysis when fine-tuned, and Large Language Models (LLM) like GPT \citep{radford-2018-gpt} are strong zero-shot and few-shot learners \citep{kheiri-2023-sentimentgpt}.

Recent work has explored collaboration between these models, such as using GPT to augment data of minority classes before fine-tuning with RoBERTa \citep{kokshun-2023-intertwining}, using GPT for aspect extraction and RoBERTa for sentiment scoring \citep{qian-2024-experience}, and escalating to LLMs when RoBERTa classification confidence was low \citep{andrade-2024-strategy}. However, leveraging external knowledge of sentiment from fine-tuned encoders to enhance LLMs remains under-explored.

This research investigates collaborative approaches between ELECTRA and GPT-4o models \citep{openai-2024b-gpt4o, openai-2024c-gpt4omini} for three-way sentiment classification (negative, neutral, positive) of reviews. Our research focused on the following hypotheses: Providing predictions from a fine-tuned ELECTRA as context to a GPT model will improve classification performance (\textbf{H1}). The improvement in performance will be less for a fine-tuned GPT (\textbf{H2}). The format of predictions in the prompt will affect performance (\textbf{H3}). Including similar examples in the prompt will improve performance (\textbf{H4}).

These hypotheses build on ELECTRA's strength in capturing nuanced sentiment patterns when fine-tuned \citep{clark-2020-electra, potts-2021-dynasent, mala-2023-efficacy}, and GPT's versatility through in-context learning \citep{radford-2019-multitask, liu-2019-roberta, kocon-2023-chatgpt, openai-2024a-gpt4}---they can perform well across diverse tasks when given the appropriate context through prompting \citep{liu-2023-prompting, khattab-2024-dspy}. Although they may struggle with emotion and nuance \citep{kocon-2023-chatgpt},  retrieved examples can improve performance \citep{zhang-2023-rag}.

To test these hypotheses, we established four baselines and conducted 28 experiments across three sentiment classification datasets: Stanford Sentiment Treebank (SST), and DynaSent Rounds 1 and 2. We used ELECTRA Base/Large and GPT-4o/4o-mini, each of which were fine-tuned (FT) on a merge of SST and DynaSent reviews.

We investigated the effects of different prompt augmentation scenarios using Declarative Self-improving Python (DSPy) \citep{khattab-2024-dspy}, a framework for programming language models. We started with a prompt to classify a review, and augmented it with knowledge from ELECTRA in the form of: the predicted class label, the probabilities of each class, similar reviews with their class labels, and combinations. We evaluated classification performance with the macro average F1 score, and cost-effectiveness by dividing total fine-tuning costs by the F1. Our key insights are the following.

\textbf{Sharing predictions boosted performance.} Augmenting GPT-4o-mini (not fine-tuned) with predictions from ELECTRA Base FT significantly improved performance over either model alone. It also yielded the lowest cost/performance ratio.

\textbf{Adding probabilities or examples did not help.} Using probabilities, or including few-shot examples, did not improve performance more than the predicted label alone for both GPT models.

\textbf{Fine-tuned GPTs performed best.} GPT-4o FT-M alone achieved the highest overall performance on the merged test set, with GPT-4o-mini FT closely following at significantly lower cost.

\textbf{Sharing predictions hurt fine-tuned GPTs.} When GPT models were fine-tuned, including ELECTRA predictions decreased performance---even when fine-tuned with the same inference-time prompt that included the ELECTRA prediction. Fine-tuning with the prediction for more epochs allowed GPT to discriminate better. Performance improved, but the cost grew significantly.

\textbf{Fine-tuned ELECTRA Large outperformed base GPTs.} ELECTRA Large fine-tuned was the best performing encoder model, and was better than both GPT-4o and GPT-4o-mini base models.

These findings offer affordable options for projects with limited resources. If fine-tuning via API is an option, a fine-tuned GPT-4o-mini is nearly as good as GPT-4o FT at 76\% less cost. Alternatively, augmenting LLM prompts with predictions from fine-tuned encoder models is an efficient way to boost performance.  For projects that want to stay local, a fine-tuned ELECTRA Large model is quite capable, and better than default GPTs.

The key contributions of this research are:
\begin{itemize}
    \item Proposes a novel collaboration where fine-tuned bidirectional encoders assist GPT models with the task of sentiment classification.
    \item Demonstrates that augmenting GPT prompts (not fine-tuned) with predictions from fine-tuned encoders significantly improves classification performance and reduces costs, achieving the lowest cost/performance ratio.
    \item Evaluates various formats for incorporating encoder output into GPT prompts, and offers practical guidelines to maximize performance.
\end{itemize}

\section{Prior Literature}

\subsection{MLMs and ELECTRA}

Masked Language Models (MLM) like BERT (Bidirectional Encoder Representations from Transformers) \citep{devlin-2019-bert} employed bidirectional encoding to obtain holistic representations of text. RoBERTa (Robustly Optimized BERT Pretraining Approach) \citep{liu-2019-roberta} optimized the pre-training approach, but both models were inefficient because learning only occurred in about 15\% of the tokens that were masked.

This led to the development of ELECTRA (Efficiently Learning an Encoder that Classifies Token Replacements Accurately) \citep{clark-2020-electra}. ELECTRA was pre-trained with two models using replaced token detection. As a result, it learned from all tokens and had comparable or better performance in a variety of tasks with less compute.

ELECTRA was found to be a top performer in sentiment classification on datasets such as SST \citep{clark-2020-electra}, DynaSent \citep{potts-2021-dynasent}, and IMDB movie reviews \citep{mala-2023-efficacy}. It was also found to be better suited for prompt-based learning due to its use of a discrimnator \citep{xia-2022-prompting}. We chose to use ELECTRA for these reasons, in addition to observing a performance gain relative to RoBERTa in early trials.

\subsection{GPT Models}

\begin{table*}[ht]
    \centering
    \setlength{\tabcolsep}{12pt}
    \small
    \caption{Examples of Merged Training Dataset}
    \begin{tabular}{l p{8cm} l l}
    \toprule
    Index & Sentence & Label & Source \\
    \midrule
    0 & Those 2 drinks are part of the HK culture and has years of history. It is so bad. & negative & dynasent\_r2 \\
    1 & I was told by the repair company that was doing the car repair that fixing the rim was "impossible" and to replace it. & negative & dynasent\_r1 \\
    2 & It is there to give them a good time . & neutral & sst\_local \\
    3 & Like leafing through an album of photos accompanied by the sketchiest of captions . & negative & sst\_local \\
    4 & Johnny was a talker and liked to have fun. & positive & dynasent\_r1 \\
    \bottomrule
    \end{tabular}
    \label{tab:merged_examples}
\end{table*}

Bidrectional transformers seemed to have an edge over early autoregressive models like GPT \citep{radford-2018-gpt} for sentiment analysis. But that edge is being whittled away by the successors of GPT pre-trained at a massive scale: GPT-3, GPT-3.5, GPT-4, and GPT-4o \citep{openai-2024a-gpt4, openai-2024b-gpt4o, openai-2024c-gpt4omini}.

For sentiment analysis of social media posts, \citet{kheiri-2023-sentimentgpt} found that GPT models significantly outperformed a number of prior models on the SemEval 2017 dataset. In contrast, \citet{kocon-2023-chatgpt} found that, although ChatGPT is versatile and competent across a wide range of tasks, it did not perform as well as RoBERTa---especially for pragmatic tasks involving detection of emotional and contextual nuances. They propose that fine-tuning ChatGPT may be necessary, which we explore in this research.

\subsection{Collaborative Approaches}

Recent work has revealed several promising approaches for collaboration between these models.

\citet{kokshun-2023-intertwining} explored a unique framework that chains GPT and RoBERTa for emotion detection. They used GPT's generative capabilities to augment training data for minority classes. The augmented dataset is then used to fine-tune RoBERTa on emotion detection.

\citet{qian-2024-experience} tapped the strengths of different models in a Natural Language Processing (NLP) pipeline to analyze stadium reviews. One GPT-3.5 model was fine-tuned to extract experience aspects, while another classified these aspects into categories. A RoBERTa model then performed sentiment scoring on the extracted aspects. We are chaining ELECTRA and GPT-4o in a similar manner here, but in a different order.

\citet{andrade-2024-strategy} investigated the benefits of collaboration between MLMs and open LLMs for sentiment classification, similar to the current research. In their ``Call-My-Big-Sibling'' (CMBS) approach, the initial classification is done with a calibrated RoBERTa model. If RoBERTa has low confidence on the classification, an open LLM like Llama 2 \citep{touvron-2023-llama2} is invoked to perform the classification task instead.

In CMBS, the final prediction is either made by RoBERTa or Llama 2---it's a decision tree. In contrast, our approach always passes the ELECTRA prediction to the LLM. If we had to come up with a similar analogy, it would be ``Show-Me-Your-Answers'' (SMYA). And then it's up to the LLM to decide if it follows the ELECTRA prediction, or decides to classify the review differently.

Most recently, \citet{charpentier-2024-gptbertboth} created GPT-BERT, a hybrid model that learns bidirectional representations like an MLM, but is also generative like a GPT. By shifting the prediction of masked tokens one position to the right, GPT-BERT can be trained on both MLM and autoregressive objectives without changing architecture. In the BabyLM Challenge 2024 benchmark, it outperformed models trained on only one objective, showing there is potential in this combined approach.

\begin{table}[ht]
    \centering
    \setlength{\tabcolsep}{12pt}
    \small
    \caption{Label Distribution for the Merged Dataset}
    \begin{tabular}{l rrr}
    \toprule
    Split & Negative & Neutral & Positive \\
    \midrule
    Train & 21,910 & 49,148 & 31,039 \\
    Validation & 1,868 & 1,669 & 1,884 \\
    Test & 2,352 & 1,829 & 2,349 \\
    \bottomrule
    \end{tabular}
    \label{tab:merged_distribution}
\end{table}

\begin{table}[ht]
    \centering
    \setlength{\tabcolsep}{12pt}
    \small
    \caption{Contribution of Sources to the Merged Dataset}
    \begin{tabular}{l r r}
    \toprule
    Dataset & Samples & Percent (\%) \\
    \midrule
    DynaSent R1 Train & 80,488 & 78.83 \\
    DynaSent R2 Train & 13,065 & 12.80 \\
    SST-3 Train & 8,544 & 8.37 \\
    \midrule
    Total & 102,097 & 100.00 \\
    \bottomrule
    \end{tabular}
    \label{tab:merged_contributions}
\end{table}

\begin{table*}[ht!]
    \centering
    \setlength{\tabcolsep}{12pt}
    \small
    \caption{Models Used in Research}
    \begin{tabular}{l l l l}
    \toprule
    Model & Provider & Access & Identifier \\
    \midrule
    ELECTRA Base & Hugging Face & Local &  google/electra-base-discriminator \\
    ELECTRA Large & Hugging Face & Local & google/electra-large-discriminator \\
    GPT-4o & OpenAI & API & gpt-4o-2024-08-06 \\
    GPT-4o-mini & OpenAI & API & gpt-4o-mini-2024-07-18 \\
    \bottomrule
    \end{tabular}
    \label{tab:model_ids}
\end{table*}

\section{Data}

Models were trained and evaluated in English on a merge of movie reviews from the Stanford Sentiment Treebank (SST) \citep{socher-2013-sst} and business reviews from DynaSent Rounds 1 and 2 \citep{potts-2021-dynasent}, licensed under Apache 2.0 and Creative Commons Attribution 4.0 respectively. See Table \ref{tab:merged_examples} for examples. By default, SST is a five-way classification (positive, somewhat positive, neutral, somewhat negative, negative). The positive and negative classes were combined to produce SST-3 (positive, neutral, negative). 

The SST-3, DynaSent R1, and DynaSent R2 datasets were randomly mixed to form a Merged dataset with 102,097 Train examples, 5,421 Validation examples, and 6,530 Test examples. See Table \ref{tab:merged_distribution} for the distribution of labels, and Table \ref{tab:merged_contributions} for a breakdown of sources. It’s worth noting that the source datasets all have class imbalances. Merging the data helps mitigate this imbalance, but there is still a majority of neutral examples in the training split. Another potential issue is that the models will learn the dominant dataset, which is DynaSent R1. As a test, the minority classes were over-sampled to create a new balanced dataset. When this was evaluated, the performance did not improve. 

\section{Models}

Four models were fine-tuned and evaluated in this research, both individually and in collaboration with each other: ELECTRA Base and Large, and GPT-4o and 4o-mini. See Table \ref{tab:model_ids} for details.

ELECTRA \citep{clark-2020-electra}, released with an Apache 2.0 license, was chosen as the bidirectional transformer because its pre-training architecture gives it an advantage over MLMs. It also outperformed RoBERTa in early trials. We evaluated both the Base (110M parameters) and Large (335M parameters) variants.

To function as a classifier, ELECTRA's output is sent through a mean pooling layer. A classifier head is appended with 2 hidden layers of dimension 1024, and a final output dimension of 3. Swish GLU \citep{shazeer-2020-gluvariant} was used as the hidden activation function, and dropout layers were added with a rate of 0.3. See Appendix \ref{app:finetune} for more details on the model architecture and hyper-parameters.

For comparison and collaboration, two GPT models were used via OpenAI's API: GPT-4o \citep{openai-2024b-gpt4o} and GPT-4o-mini \citep{openai-2024c-gpt4omini}. Although the full specifications are not public, they are state-of-the-art autoregressive language models with strong zero-shot capabilities. GPT-4o is described as a ``high-intelligence flagship model for complex, multi-step tasks.'' GPT-4o-mini is described as an ``affordable and intelligent small model for fast, lightweight tasks.''

\section{Methods}

Our research progressed through the following stages. Code and datasets are available at: \url{https://github.com/jbeno/sentiment}.

\subsection{ELECTRA Baseline \& Fine-tuning}

We first developed a training pipeline to support interactivity and distributed training across multiple GPUs. Training progress was tracked through Weights and Biases so we could monitor train/validation metrics (loss, macro F1, accuracy) across epochs. The final models were selected from checkpoints at convergence, or just before train/validation metrics started to diverge.

Two baseline models were established by training only classifier heads for ELECTRA Base and Large. Hyper-parameters were consistent with the fully fine-tuned versions. The fine-tuning process involved a number of trials on Lambda Labs multi-GPU instances to identify the best hyper-parameters, optimizer, and learning rate schedule. See Appendix \ref{app:finetune} for the final configuration.

We also explored alternative approaches including an ensemble of binary classifiers, and additional fine-tuning on DynaSent R2 and SST-3, but these did not outperform our initial approach. 

\subsection{GPT Data Preparation \& Fine-tuning}

To use OpenAI's fine-tuning API, we converted the Merged training data to JSONL format that defined the System, User, and Assistant roles. We noticed that if the context at inference time varied even slightly from the fine-tuning context, performance would suffer. So we created three templates to enable better comparisons between fine-tuned and default models using the same DSPy signatures (see Appendix \ref{app:ft_templates}):

\begin{itemize}
    \item \textbf{Minimal (FT-M): }No prompt other than System role. User role only contained the review sentence.
    \item \textbf{Prompt (FT): }Default fine-tuning. User role included full DSPy prompt.
    \item \textbf{Prompt with Label (FT-L): }User role included DSPy prompt with ELECTRA predicted label.
\end{itemize}

We included the ELECTRA predictions in the third template to align the fine-tuning context with the inference time context, but also to provide an opportunity for the GPT models to learn from the ELECTRA predictions.  In total there were 9 fine-tuning jobs (see Table \ref{tab:gpt-finetuning}, and Appendix \ref{app:finetune2} for GPT fine-tuning details).

\begin{table}[h]
\small
\centering
\caption{Fine-Tuning Job Details}
\begin{tabular}{llll}
\toprule
Model & Code & Format & Epochs \\
\midrule
4o-mini & FT-M & Minimal & 1 \\
4o-mini & FT & Prompt & 1 \\
4o-mini & FT-L & Prompt w/Base Label & 1 \\
4o-mini & FT-L & Prompt w/Base Label & 5 \\
4o-mini & FT-L & Prompt w/Large Label & 1 \\
4o-mini & FT-L & Prompt w/Large Label & 5 \\
4o & FT-M & Minimal & 1 \\
4o & FT & Prompt & 1 \\
4o & FT-L & Prompt w/Large Label & 1 \\
\bottomrule
\end{tabular}
\label{tab:gpt-finetuning}
\end{table}

\subsection{DSPy Signatures \& Modules}

With DSPy, you create modules (ex: Classify, ClassifyWithExamples), signatures (input/output templates, ex: review + examples $\rightarrow$ classification), define metrics (ex: classification\_match) and evaluators of data, and use optimizers to find the best performing prompt or module parameters.

We explored a variety of approaches to integrating ELECTRA's output into GPT's decision-making process. Each approach was implemented as a custom DSPy signature and module (see Appendix \ref{app:prompts} for the full examples).

\textbf{Classification Prompt.} Prompt to ``Classify the sentiment of a review as either `negative', `neutral', or `positive'.'' One input field `review' described as ``The review text to classify.'' and one output field `classification' described as ``One word representing the sentiment classification: `negative', `neutral', or `positive' (do not repeat the field name, do not use `mixed')''.

\textbf{Predicted Label.} Classification prompt with an additional input field `classifier\_decision' described as ``The sentiment classification proposed by a model fine-tuned on sentiment.'' During evaluation, the DSPy module first sends the review through the ELECTRA model to obtain its prediction. This output is then inserted into the signature.

\textbf{Probabilities.} Classification prompt, but instead of `classifier\_decision' it featured three input fields for the probabilities of each class as obtained from the ELECTRA model. For example: `negative\_probability' was described as ``Probability the review is negative from a model fine-tuned on sentiment''. The float is converted to a percent to make it easier for the model to interpret.

\textbf{Prediction \& Probabilities.} Same as Probabilities, but it also included the `classifier\_decision' to emphasize the final decision made by ELECTRA. 

\textbf{Top Examples.} A custom retriever was created from 300 reviews in the Validation split. During inference, input text is run through the fine-tuned ELECTRA Large model to extract the output representations (prior to the classifier head). The top five matches and class labels based on cosine similarity are shown as few-shot examples. This signature had `classifier\_decision' plus an `examples' field described as ``A list of examples that demonstrate different sentiment classes.''

\textbf{Balanced Examples.} If ELECTRA was wrong, and the Top Examples were all of the same class, it might be hard for GPT to make an independent decision. To compensate, in Balanced Examples, a different retriever was used that retrieved a total of six examples (the top two examples from each class) to ensure the few-shot examples with true labels did not bias the answer toward a particular class---although that might be desirable sometimes.  

\textbf{All of the Above.} And lastly, a final DSPy signature had all of the above context from ELECTRA included: classification prompt, predicted label, probabilities, and top five examples (not balanced). It was unclear if providing all this information would help GPT make a decision, or if the large number of tokens would dilute the signal.

We then conducted two of the four baselines, and 26 of the 28 experiments (see Table \ref{tab:experiment_summary}) using these DSPy signatures and modules. The fine-tuned ELECTRA models and retriever were instantiated locally for inference, and the GPT models were accessed via OpenaAI API. To address single-run concerns, each baseline and experiment was run a second time with a different random seed (123 vs. 42) and temperature (0.1 vs. 0.0).

\section{Results}

\begin{table*}[ht]
    \begin{threeparttable}
    \centering
    \setlength{\tabcolsep}{3pt}
    \scriptsize
    \caption{Summary of Model Configuration, Test Set Performance, and Cost}
    \begin{tabular}{l l l l r r r r r r}
    \toprule
    & & & & \multicolumn{1}{c}{Merged\tnote{3}} & \multicolumn{1}{c}{DynaSent R1} & \multicolumn{1}{c}{DynaSent R2} & \multicolumn{1}{c}{SST-3} & \multicolumn{2}{c}{Cost (\$)\tnote{5}} \\
    \cmidrule(lr){5-5} \cmidrule(lr){6-6} \cmidrule(lr){7-7} \cmidrule(lr){8-8} \cmidrule(lr){9-10}
    ID\tnote{1} & GPT\tnote{2} & ELECTRA & Description & Macro F1\tnote{4} & Macro F1\tnote{4} & Macro F1\tnote{4} & Macro F1\tnote{4} & FT & /F1 \\
    \midrule
      B1 & --- & Base & Baseline, Classifier head & 69.51 ± 0.20 & 70.86 ± 0.15 & 61.39 ± 0.28 & 60.60 ± 0.36 & \cellcolor{lightgray}0.65 & \cellcolor{lightgray}0.01 \\
    B2 & --- & Large & Baseline, Classifier head & 67.94 ± 0.08 & 69.70 ± 0.04 & 59.78 ± 0.00 & 57.95 ± 0.37 & 2.51 & 0.04 \\
    B3 & 4o-mini & --- & Baseline (Zero shot) & $^\dagger$79.41 ± 0.16 & \cellcolor{lightgray}81.16 ± 0.05 & 77.02 ± 0.47 & 69.99 ± 0.97 & --- & --- \\
    B4 & 4o & --- & Baseline (Zero shot) & \cellcolor{lightgray}79.97 ± 0.24 & 80.95 ± 0.25 & \cellcolor{lightgray}80.14 ± 0.12 & \cellcolor{lightgray}72.08 ± 0.17 & --- & --- \\
    \midrule
      E1 & --- & Base FT & Fine-tune all layers & $^\dagger$79.14 ± 0.22 & 82.12 ± 0.02 & 70.67 ± 1.64 & 69.04 ± 1.29 & \cellcolor{lightgray} 9.73 & \cellcolor{lightgray} 0.12 \\
    E2 & --- & Large FT & Fine-tune all layers & \cellcolor{lightgray}82.76 ± 0.57 & \cellcolor{lightgray}86.22 ± 0.44 & \cellcolor{lightgray}77.33 ± 1.46 & \cellcolor{lightgray}71.77 ± 1.22 & 53.26 & 0.65 \\
    \midrule
      E3 & 4o-mini & Base FT & Prompt, Label & $^\dagger$82.50 ± 0.34 & 86.40 ± 0.15 & 75.33 ± 1.22 & 70.88 ± 1.20 & \cellcolor{lightgray} \textbf{9.73} & \cellcolor{lightgray} \textbf{0.12} \\
    E4 & 4o-mini & Large FT & Prompt, Label & \cellcolor{lightgray}83.80 ± 0.43 & \cellcolor{lightgray}87.71 ± 0.27 & 78.73 ± 1.12 & 71.77 ± 1.10 & 53.26 & 0.64 \\
    E5 & 4o-mini & Large FT & Prompt, Label, Examples (Few shot) & 83.42 ± 0.30 & 86.94 ± 0.28 & 79.50 ± 1.12 & \cellcolor{lightgray}72.33 ± 0.49 & 53.26 & 0.64 \\
    E6 & 4o-mini & Large FT & Prompt, Label, Balanced Ex. (Few shot) & 82.98 ± 0.42 & 86.28 ± 0.62 & 79.87 ± 0.36 & 71.98 ± 0.75 & 53.26 & 0.64 \\
    E7 & 4o-mini & Large FT & Prompt, Probs & 83.27 ± 0.37 & 86.60 ± 0.23 & 79.41 ± 0.69 & 72.26 ± 1.03 & 53.26 & 0.64 \\
    E8 & 4o-mini & Large FT & Prompt, Label, Probs & 83.66 ± 0.32 & 87.22 ± 0.28 & \cellcolor{lightgray}79.98 ± 0.36 & 71.78 ± 1.06 & 53.26 & 0.64 \\
    E9 & 4o-mini & Large FT & Prompt, Label, Probs, Examples & 83.19 ± 0.39 & 86.58 ± 0.60 & 78.99 ± 0.42 & 71.94 ± 0.64 & 53.26 & 0.64 \\
    \midrule
      E10 & 4o-mini FT & --- & Fine-tune w/prompt & \cellcolor{lightgray}86.70 ± 0.11 & \cellcolor{lightgray}89.65 ± 0.30 & 87.00 ± 0.13 & \cellcolor{lightgray}\textbf{75.83 ± 0.21} & 33.15 & 0.38 \\
    E11 & 4o-mini FT 5 & --- & Fine-tune w/prompt (5 epochs) & 84.86 ± 0.13 & 87.74 ± 0.13 & 86.22 ± 0.40 & 75.38 ± 0.32 & 165.75 & 1.95 \\
    E12 & 4o-mini FT-M & --- & Minimal fine-tune & 86.51 ± 0.06 & 89.57 ± 0.18 & \cellcolor{lightgray}87.13 ± 0.22 & 75.74 ± 0.17 & \cellcolor{lightgray}16.60 & \cellcolor{lightgray}0.19 \\
    E13 & 4o-mini FT & Base FT & Prompt, Label, FT w/prompt & 81.06 ± 0.52 & 84.67 ± 0.14 & 73.06 ± 2.03 & 69.70 ± 1.77 & 42.88 & 0.53 \\
    E14 & 4o-mini FT-L & Base FT & Prompt, Label, FT w/prompt, label & 81.84 ± 0.26 & 85.20 ± 0.06 & 77.29 ± 1.22 & 70.70 ± 1.39 & 49.31 & 0.60 \\
    E15 & 4o-mini FT-L 5 & Base FT & Prompt, Label, FT w/prompt, label (5 epochs) & 83.67 ± 0.30 & 86.38 ± 0.51 & 81.19 ± 0.62 & 75.02 ± 0.03 & 207.64 & 2.48 \\
    E16 & 4o-mini FT & Large FT & Fine-tune w/prompt & 83.94 ± 0.09 & 87.57 ± 0.11 & 80.17 ± 0.28 & 72.46 ± 0.00 & 86.41 & 1.03 \\
    E17 & 4o-mini FT-L & Large FT & Fine-tune w/prompt, label & 84.12 ± 0.06 & 87.58 ± 0.10 & 80.75 ± 0.20 & 73.34 ± 0.06 & 92.84 & 1.10 \\
    E18 & 4o-mini FT-L 5 & Large FT & Fine-tune w/prompt, label (5 epochs) & 84.83 ± 0.06 & 87.75 ± 0.17 & 84.37 ± 0.81 & 75.59 ± 0.01 & 251.17 & 2.96 \\
    \midrule
      E19 & 4o & Large FT & Prompt, Label & 83.19 ± 0.01 & 85.71 ± 0.00 & \cellcolor{lightgray}82.06 ± 0.11 & \cellcolor{lightgray}73.48 ± 0.06 & \cellcolor{lightgray}53.26 & \cellcolor{lightgray}0.64 \\
    E20 & 4o & Large FT & Prompt, Label, Examples (Few shot) & 83.29 ± 0.28 & 86.11 ± 0.14 & 81.48 ± 0.07 & 72.96 ± 1.27 & \cellcolor{lightgray}53.26 & \cellcolor{lightgray}0.64 \\
    E21 & 4o & Large FT & Prompt, Label, Balanced Ex. (Few shot) & 83.19 ± 0.28 & 86.01 ± 0.19 & 81.04 ± 0.21 & 72.88 ± 1.03 & \cellcolor{lightgray}53.26 & \cellcolor{lightgray}0.64 \\
    E22 & 4o & Large FT & Prompt, Probs & 82.99 ± 0.47 & 86.37 ± 0.45 & 78.42 ± 1.05 & 71.90 ± 1.04 & \cellcolor{lightgray}53.26 & \cellcolor{lightgray}0.64 \\
    E23 & 4o & Large FT & Prompt, Label, Probs & \cellcolor{lightgray}83.31 ± 0.33 & \cellcolor{lightgray}86.69 ± 0.35 & 79.46 ± 0.33 & 72.17 ± 0.97 & \cellcolor{lightgray}53.26 & \cellcolor{lightgray}0.64 \\
    E24 & 4o & Large FT & Prompt, Label, Probs, Examples & 83.04 ± 0.42 & 86.53 ± 0.29 & 78.47 ± 1.00 & 71.83 ± 1.20 & \cellcolor{lightgray}53.26 & \cellcolor{lightgray}0.64 \\
    \midrule
      E25 & 4o FT & --- & Fine-tune w/prompt & 86.79 ± 0.06 & 90.46 ± 0.03 & 88.14 ± 0.28 & 73.09 ± 0.01 & 276.24 & 3.18 \\
    E26 & 4o FT-M & --- & Minimal fine-tune & \cellcolor{lightgray}\textbf{86.99 ± 0.00} & \cellcolor{lightgray}\textbf{90.57 ± 0.00} & \cellcolor{lightgray}\textbf{89.00 ± 0.00} & \cellcolor{lightgray}73.99 ± 0.00 & \cellcolor{lightgray}138.37 & \cellcolor{lightgray}1.59 \\
    E27 & 4o FT & Large FT & Fine-tune w/prompt & 84.03 ± 0.30 & 87.90 ± 0.13 & 80.01 ± 0.73 & 72.00 ± 1.15 & 329.50 & 3.93 \\
    E28 & 4o FT-L & Large FT & Fine-tune w/prompt, label & 84.37 ± 0.19 & 87.81 ± 0.09 & 81.28 ± 1.03 & 73.10 ± 0.66 & 383.10 & 4.55 \\
    \bottomrule
    \end{tabular}
    
    \begin{tablenotes}[flushleft]
    \scriptsize
    \item \textbf{Bold} = best overall, \colorbox{lightgray}{highlighted} = best in section 
    \item$^\dagger$Scores relevant to Hypothesis 1 (ELECTRA prediction improving non-fine-tuned GPT performance)
    \item\textsuperscript{1} Some ID numbers changed from their original ID in the research repo.
    \item\textsuperscript{2} GPT fine-tuning types: FT = fine-tune with prompt, FT-M = minimal without prompt, FT-L = with prompt including ELECTRA label, FT 5 = 5 epochs vs. 1
    \item\textsuperscript{3} Merged dataset: Combination of test splits from DynaSent R1/R2 and SST-3
    \item\textsuperscript{4} Each experiment was run twice with different random seeds (42, 123) and temperature (0.0, 0.1); values reported are means ± standard deviations. Standard deviations are based on two runs (n=2) and should be interpreted with caution.
    \item\textsuperscript{5} Cost: FT = Fine-tuning cost, no inference-time API charges. Ratio is FT cost divided by F1 score.
    \end{tablenotes}
    
    \label{tab:experiment_summary}
    \end{threeparttable}
\end{table*}

Our experiments revealed significant differences in performance across baseline, fine-tuning, and collaborative scenarios. See Table \ref{tab:experiment_summary} for the mean macro average F1 between the two runs. Appendix \ref{app:raw_results} has the raw data of each run.

\textbf{Baselines.} Regarding baselines, both GPT models outperformed the ELECTRA classifiers, with GPT-4o achieving a 79.97 mean macro F1 and GPT-4o-mini scoring 79.41, compared to ELECTRA Base (69.51) and Large (67.94). This demonstrates the strong zero-shot capabilities of GPT models.

\begin{figure*}[ht]
    \centering
    \includegraphics[width=\textwidth]{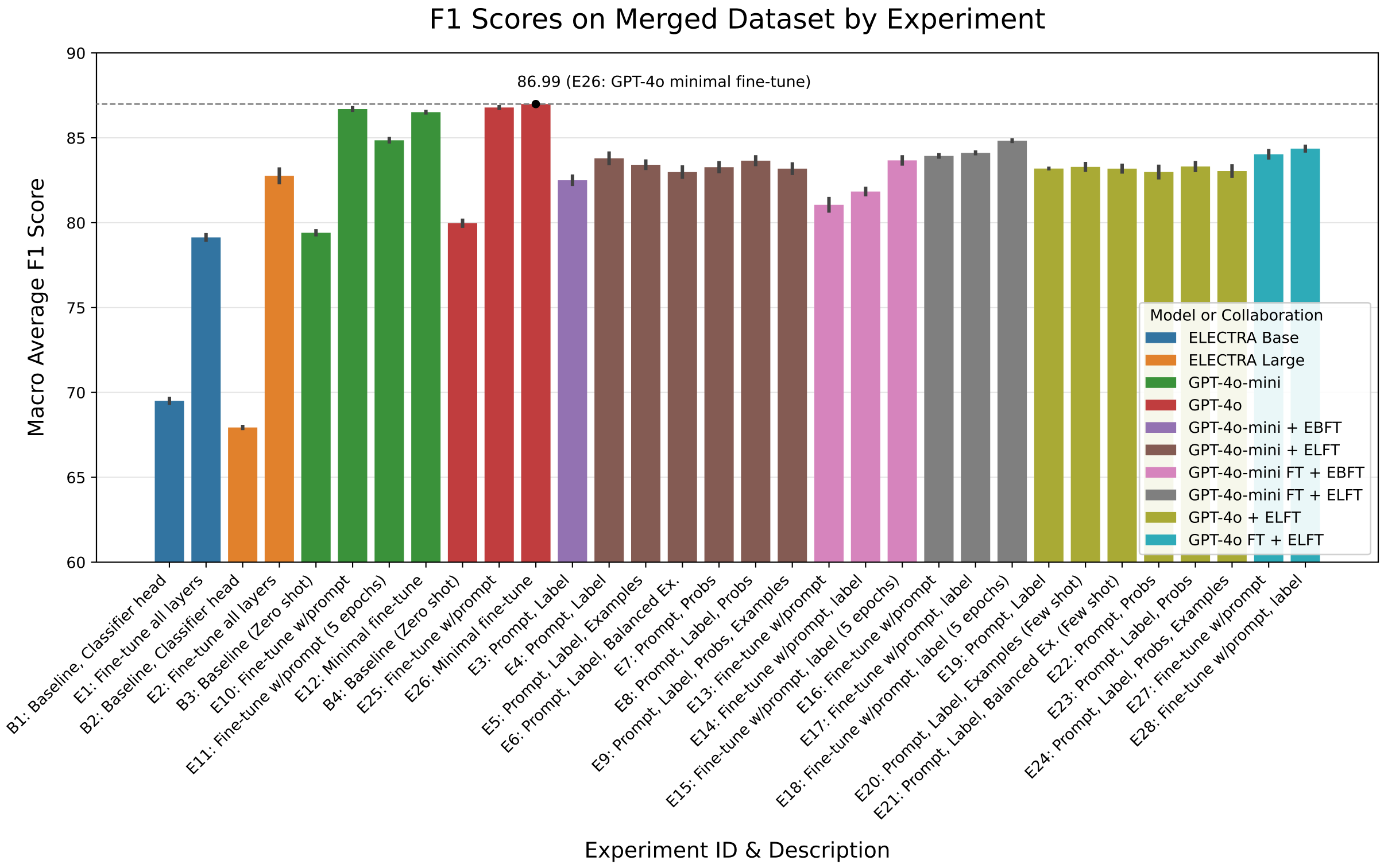}
    \caption{Mean Macro F1 Scores on Merged Dataset by Experiment}
    \label{fig:f1_scores_by_experiment}
\end{figure*}

\textbf{Fine-tuning.} Fine-tuning improved performance across all models. ELECTRA Base's mean macro F1 increased from 69.51 to 79.14, while ELECTRA Large showed greater gains, improving from 67.94 to 82.76. This improvement is the result of fine-tuning all layers---the baselines had the same classifier head. Fine-tuned GPT models had the highest scores (see Figure \ref{fig:f1_scores_by_experiment}), with GPT-4o-mini FT rising from 79.41 to 86.70, and GPT-4o FT-M achieving 86.99 with the minimal template.

\begin{figure}[h]
    \centering
    \includegraphics[width=\columnwidth]{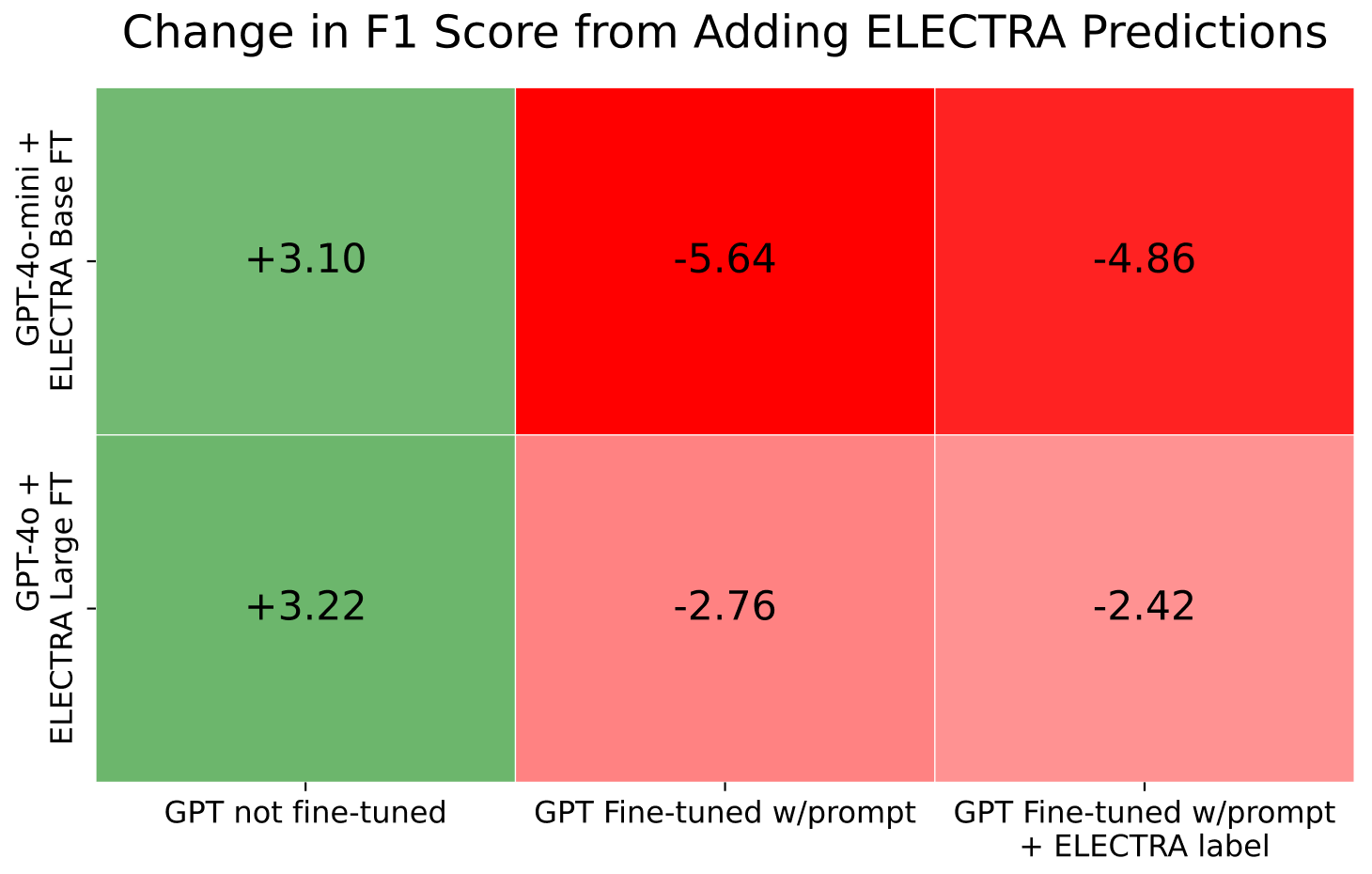}
    \caption{Change in Mean F1 from Adding Predictions}
    \label{fig:ft_score_differences_heatmap}
\end{figure}

\textbf{Sharing Predictions.} The effect of adding ELECTRA predictions to GPT prompts depended on if the GPT model was fine-tuned (see Figure \ref{fig:ft_score_differences_heatmap} for the differences in mean F1). Sharing ELECTRA Base predictions with GPT-4o-mini (not fine-tuned) significantly improved the macro F1 in round one from 79.52 to 82.74 (p < 0.0001, McNemar's test and bootstrap analysis), a +3.22 gain. There was an even greater gain of +3.97 points when ELECTRA Large predictions were shared (from 79.52 to 83.49, p < 0.0001). Similarly, including ELECTRA Large predictions with GPT-4o improved the macro F1 from 80.14 to 83.18 (p < 0.001) in round one, a +3.04 gain.

\begin{figure}[ht]
    \centering
    \includegraphics[width=\columnwidth]{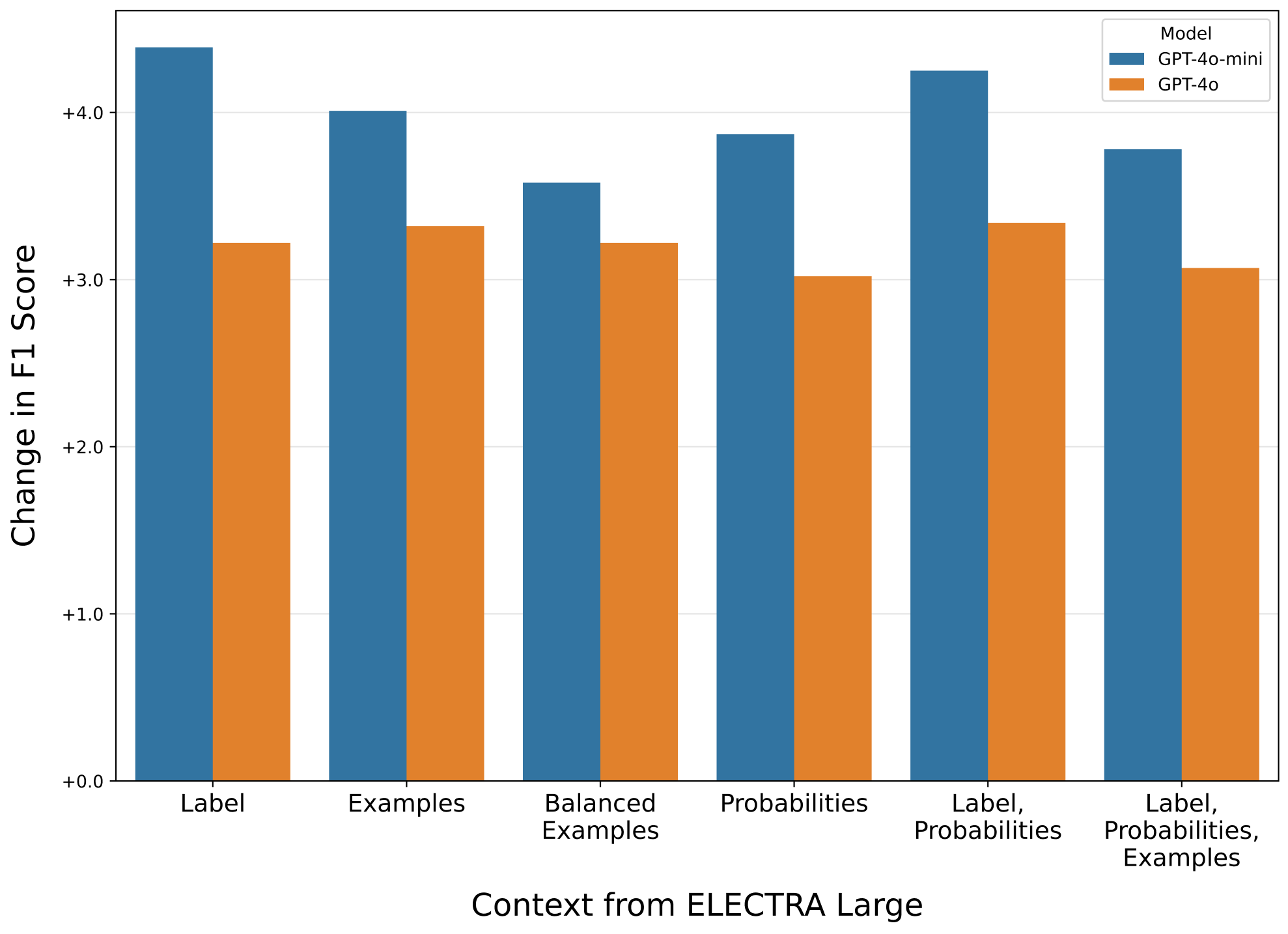}
    \caption{Impact of Context on Mean F1 Score}
    \label{fig:context_impact_both}
\end{figure}

However, sharing ELECTRA predictions with fine-tuned GPT models actually decreased performance. GPT-4o-mini FT's mean macro F1 dropped from 86.70 to 81.06 when including ELECTRA Base predictions, and to 81.84 when fine-tuned with the predictions included in the prompt. Similarly, GPT-4o FT's mean F1 fell from 86.79 to 84.03 when including ELECTRA Large predictions, and to 84.37 when fine-tuned with them.

\textbf{Few-shot Examples.} Some contexts performed better than others for specific model combinations (see Figure \ref{fig:context_impact_both}). Providing few-shot examples in addition to the predicted label was mostly the same or worse than using the label alone. However, when looking at the more challenging DynaSent Round 2 dataset, GPT-4o-mini saw some benefit. Including just the ELECTRA Large predicted label produced a mean macro F1 of 78.73. Adding examples increased the mean macro F1 to 79.50 (+0.77), and balanced examples increased it to 79.87 (+1.14). 

\textbf{Sharing Probabilities.} Using probabilities instead of (or in addition to) the predicted label was mostly the same or worse than using the label alone. However, similar to using examples, the more challenging datasets saw some benefit. For DynaSent Round 2, GPT-4o-mini had a mean macro F1 of 78.73 with just the ELECTRA Large predicted label. Using probabilities instead changed it to 79.41, and using the label with probabilities increased it to 79.98. A similar minor improvement was seen with SST on this dataset.

\textbf{Datasets.} Performance also varied across datasets. GPT-4o FT-M achieved the top scores on DynaSent R1 (90.57 mean macro F1) and DynaSent R2 (89.00). Surprisingly, GPT-4o-mini FT---the smaller model---performed best on SST-3 with a 75.83 mean macro F1, exceeding even GPT-4o FT's performance of 73.99.

\textbf{Cost.} The most cost-effective approach was ELECTRA Base FT with GPT-4o-mini (not fine-tuned) at \$0.12 per F1 point. GPT-4o-mini FT provided a good compromise at \$0.38 per F1 point, while GPT-4o FT-L with ELECTRA Large FT proved most expensive at \$4.55 per F1 point.

\begin{figure}[t]
    \centering
    \includegraphics[width=\columnwidth]{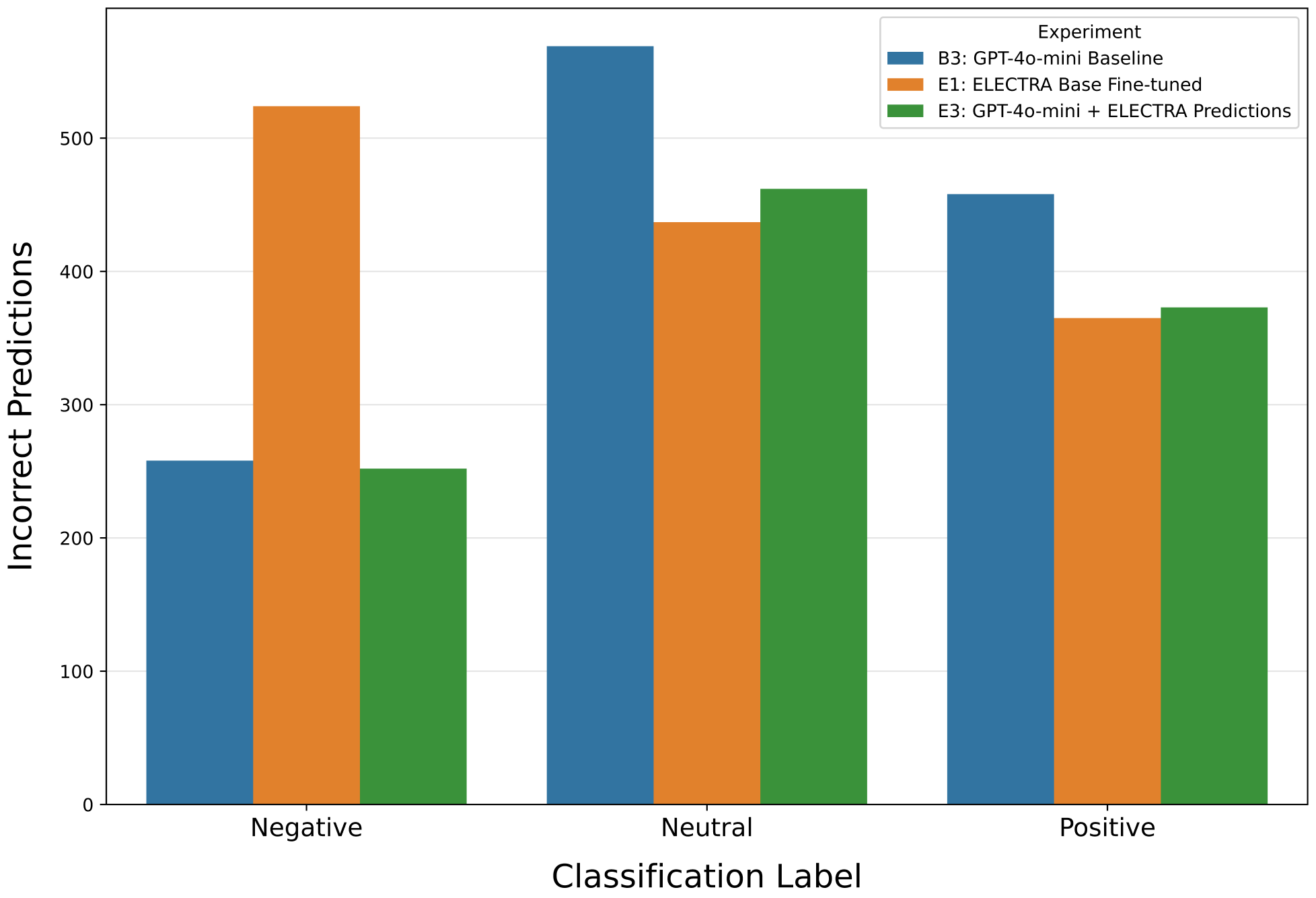}
    \caption{Incorrect Predictions by Label (Round 1)}
    \label{fig:b3_e1_e3_incorrect}
\end{figure}

\section{Analysis}

\textbf{H1. Sharing predictions would boost performance.} The significant improvement in GPT-4o-mini's performance when augmented with ELECTRA Base FT or Large FT predictions strongly supports H1. We also saw a similar boost for GPT-4o with ELECTRA Large FT predictions.

However, following ELECTRA's predictions had mixed results. When GPT-4o-mini changed its decision and followed ELECTRA Base FT in round one, it was correct 548 times and wrong 412 times (+136 net improvement, 57.08\% success rate). When GPT-4o changed its decision and followed ELECTRA Large FT, it was correct 521 times and wrong 481 times (+40 net improvement, 52\% success rate).

Most of the improvement was in the neutral and positive classes (see Figure \ref{fig:b3_e1_e3_incorrect}).  There was barely any improvement in the negative class, but importantly---it didn't worsen. ELECTRA Base FT had more than double the incorrect negative predictions, but GPT-4o-mini did not follow them. The negative class was 21.46\% of the Merged dataset, so ELECTRA may not have learned it well. Conversely, GPT-4o followed more of the negative predictions, and performance suffered. 

DynaSent R1 was the dominant source of the Merged dataset (80,488 samples, or 78.83\%), and saw the most improvement. It could be that ELECTRA learned this dataset the most, but it also represented less challenging reviews.

\textbf{H2. Improvement would be less for fine-tuned GPTs.} H2 was supported more strongly than anticipated. For a fine-tuned GPT model, including the ELECTRA prediction actually decreased performance (see Figure \ref{fig:ft_score_differences_heatmap}). Initially, we thought this was because the fine-tuning context did not include the ELECTRA prediction in the prompt. But we still saw a decrease in performance (although less) when it was included.

GPT was trained to be a helpful chat assistant that thinks through problems, evaluates information critically, and synthesizes knowledge. When presented with an ELECTRA prediction, it can decide when to follow the prediction and when to rely on its own judgment. In contrast, when fine-tuned specifically for sentiment classification, the model is optimized to map directly from input text to sentiment labels, potentially bypassing the critical thinking that made the base model effective at filtering ELECTRA's suggestions.

It may also be that it takes more time to learn the nuances of when ELECTRA is right vs. wrong. When we extended fine-tuning from 1 to 5 epochs, GPT was better able to discriminate ELECTRA's predictions. The follow rate decreased from 96.34\% to 89.66\%, and the discrimination gap---the difference between following correct and incorrect predictions---went from 16.89\% to 35.95\%.
Although additional epochs improved performance, it comes with a significant increase to the cost: \$207.64 to \$251.17 vs. only \$33.15 to just fine-tune GPT-4o-mini---with better performance.

\textbf{H3. Format of prediction would impact performance.} Although using the predicted label alone was best, GPT-4o-mini saw more variability in performance between the different contexts, as shown in Figure \ref{fig:context_impact_both}. However, for GPT-4o, there was not much of a difference when looking at the mean change in F1 score. 

\textbf{H4. Including examples would improve performance.} Contrary to this hypothesis, including examples did not improve performance for either GPT-4o-mini or GPT-4o when looking at the mean change in F1 score. It either had no effect (GPT-4o) or actually decreased performance (GPT-4o-mini).

\section{Conclusion}

This research investigated collaborative approaches to sentiment classification between bidirectional transformers and LLMs. Our results show that augmenting prompts with predictions from a fine-tuned ELECTRA can significantly improve performance when the GPT model is not fine-tuned---up to +4.39 points of gain in the mean macro F1 score. Including probabilities or similar examples improved performance slightly on the more challenging datasets. However, this collaborative benefit disappeared when the GPT models were fine-tuned.

It is possible that the fine-tuning process hinders the critical thinking that's present in the default GPT models. Fine-tuning over more epochs improves the situation---GPT gets smarter about when to follow ELECTRA vs. when to ignore it. However, this comes at a significant cost increase---six to seven times the cost of fine-tuning GPT-4o-mini alone, and the performance is still not comparable.

Our findings offer several cost-effective paths for sentiment analysis projects. For organizations that can fine-tune via API, GPT-4o-mini FT offers nearly equivalent performance to GPT-4o FT-M (86.70 vs 86.99 mean macro F1) at 76\% lower cost (\$0.38 vs \$1.59/F1 point). For those with data privacy concerns or resource constraints, GPT-4o-mini with ELECTRA Base FT had the best cost/performance ratio (\$0.12/F1 point). Projects that need to stay completely local can fine-tune ELECTRA Large, which outperformed both base GPT models.

Future work could explore optimization of inference-time prompts through DSPy, and alternate System role instructions during fine-tuning. In addition, this collaborative approach could be extended to different datasets/domains, classification tasks, and model pairings. There may also be potential for including multiple predictions from an ensemble of models. A new collaborative scenario would be fine-tuning GPTs on the ELECTRA output representations.

\section{Limitations}

The cost/performance evaluation only considered the fine-tuning costs to achieve the reported macro F1 on the test set. In practice, there may be ongoing costs for inference time API calls for hosted GPT models. The time and cost required to fine-tune different datasets will vary, and this will affect the relative cost/performance differences. 

Our research was limited to GPT-4o and GPT-4o-mini from OpenAI, but there are many GPT-style models available. Some of these are open source and can be fine-tuned locally or on hosted compute resources, such as Llama 2 or 3 from Meta \citep{touvron-2023-llama2, dubey-2024-llama}. The cost of fine-tuning an open source GPT model without API fees was not evaluated.

Resource and time constraints prevented us from exploring every possible collaborative scenario. Once we saw ELECTRA Large FT performed better than ELECTRA Base FT, we only evaluated the output from Large in the different prompt contexts for both GPT-4o and GPT-4o-mini.

Many experimental runs involved prompting the GPT models, and these prompts could be further optimized with frameworks like DSPy. Some of the prompts are verbose, and these additional tokens could dilute the signal of the ELECTRA context.

Although we quantified changes in GPT classification decisions before and after ELECTRA predictions were included, we did not have time to thoroughly explore \textit{why} the ELECTRA prediction was ignored in some cases, but followed in others. Future research could prompt the model to explain their decision, and the responses could be analyzed. This additional instruction would likely affect the classification decision---similar to Chain of Thought \citep{wei-2022-chain}---and would be a new experimental scenario.

We tried to give the GPT models a chance to learn when ELECTRA was right or wrong during fine-tuning (GPT-4o FT-L, GPT-4o-mini FT-L). There may be a more direct way of fine-tuning GPT so that it learns when to trust ELECTRA’s prediction, and when to decide for itself.

The data used for this research was a merge of movie and business reviews from SST-3 and DynaSent R1/R2. The majority of the reviews were written in English and relatively short in length. Assessing generalization to other domains, languages and longer lengths would require future research.

\section{Acknowledgments}

Thank you to the Stanford Engineering Center for Global and Online Education (CGOE) for the opportunity to conduct this research as part of the Natural Language Understanding class (XCS224U), and to Professor Christopher Potts, Insop Song, and Petra Parikova for your support and guidance. 

\bibliography{custom}

\clearpage
\onecolumn\appendix

\section{Results of Individual Runs}
\label{app:raw_results}

\begin{table}[!ht]
    \begin{threeparttable}
    \centering
    \setlength{\tabcolsep}{3pt}
    \tiny
    \caption{Round 1 Results}
    \begin{tabular}{l l l l r r r r r r r r r r}
    \toprule
    & & & & \multicolumn{2}{c}{Merged} & \multicolumn{2}{c}{DynaSent R1} & \multicolumn{2}{c}{DynaSent R2} & \multicolumn{2}{c}{SST-3} & \multicolumn{2}{c}{Cost (\$)} \\
    \cmidrule(lr){5-6} \cmidrule(lr){7-8} \cmidrule(lr){9-10} \cmidrule(lr){11-12} \cmidrule(lr){13-14}
    ID & GPT\tnote{1} & ELECTRA & Description & F1 & Acc & F1 & Acc & F1 & Acc & F1 & Acc & FT & /F1 \\
    \midrule
      B1 & --- & Base & Baseline, Classifier head & 69.65 & 69.83 & 70.96 & 71.28 & 61.59 & 61.67 & 60.85 & 70.14 & 0.65 & 0.01 \\
    B2 & --- & Large & Baseline, Classifier head & 67.88 & 68.06 & 69.72 & 70.06 & 59.78 & 59.72 & 57.68 & 67.51 & 2.51 & 0.04 \\
    B3 & 4o-mini & --- & Baseline (Zero shot) & 79.52 & 80.34 & \cellcolor{lightgray}81.12 & \cellcolor{lightgray}81.00 & 77.35 & 77.92 & 70.67 & 80.05 & --- & --- \\
    B4 & 4o & --- & Baseline (Zero shot) & \cellcolor{lightgray}80.14 & \cellcolor{lightgray}80.74 & \cellcolor{lightgray}81.12 & 80.94 & \cellcolor{lightgray}80.22 & \cellcolor{lightgray}80.56 & \cellcolor{lightgray}72.20 & \cellcolor{lightgray}80.45 & --- & --- \\
    \midrule
      E1 & --- & Base FT & Fine-tune all layers & 79.29 & 79.69 & 82.10 & 82.14 & 71.83 & 71.94 & 69.95 & 78.24 & \cellcolor{lightgray} 9.73 & \cellcolor{lightgray} 0.12 \\
    E2 & --- & Large FT & Fine-tune all layers & \cellcolor{lightgray}82.36 & \cellcolor{lightgray}82.96 & \cellcolor{lightgray}85.91 & \cellcolor{lightgray}85.83 & \cellcolor{lightgray}76.29 & \cellcolor{lightgray}76.53 & \cellcolor{lightgray}70.90 & \cellcolor{lightgray}80.36 & 53.26 & 0.65 \\
    \midrule
      E3 & 4o-mini & Base FT & Prompt, Label & 82.74 & 83.35 & 86.50 & 86.44 & 76.19 & 76.53 & 71.72 & 80.54 & \cellcolor{lightgray} \textbf{9.73} & \cellcolor{lightgray} \textbf{0.12} \\
    E4 & 4o-mini & Large FT & Prompt, Label & \cellcolor{lightgray}83.49 & \cellcolor{lightgray}84.21 & \cellcolor{lightgray}87.52 & \cellcolor{lightgray}87.47 & 77.94 & 78.47 & 70.99 & 80.77 & 53.26 & 0.64 \\
    E5 & 4o-mini & Large FT & Prompt, Label, Examples (Few shot) & 83.20 & 83.80 & 86.74 & 86.64 & 78.71 & 79.03 & \cellcolor{lightgray}71.98 & 80.72 & 53.26 & 0.64 \\
    E6 & 4o-mini & Large FT & Prompt, Label, Balanced Ex. (Few shot) & 82.68 & 83.28 & 85.84 & 85.69 & 79.61 & 80.00 & 71.45 & 80.41 & 53.26 & 0.64 \\
    E7 & 4o-mini & Large FT & Prompt, Probs & 83.01 & 83.60 & 86.44 & 86.36 & 78.92 & 79.17 & 71.53 & 80.54 & 53.26 & 0.64 \\
    E8 & 4o-mini & Large FT & Prompt, Label, Probs & 83.43 & 84.12 & 87.02 & 86.94 & \cellcolor{lightgray}79.72 & \cellcolor{lightgray}80.14 & 71.03 & 80.81 & 53.26 & 0.64 \\
    E9 & 4o-mini & Large FT & Prompt, Label, Probs, Examples & 82.91 & 83.54 & 86.15 & 86.06 & 78.69 & 79.03 & 71.49 & \cellcolor{lightgray}80.90 & 53.26 & 0.64 \\
    \midrule
      E10 & 4o-mini FT & --- & Fine-tune w/prompt & \cellcolor{lightgray}86.77 & \cellcolor{lightgray}87.26 & \cellcolor{lightgray}89.86 & \cellcolor{lightgray}89.75 & 86.90 & \cellcolor{lightgray}87.08 & \cellcolor{lightgray}\textbf{75.68} & \cellcolor{lightgray}\textbf{83.26} & 33.15 & 0.38 \\
    E11 & 4o-mini FT 5 & --- & Fine-tune w/prompt (5 epochs) & 84.95 & 85.27 & 87.83 & 87.67 & 85.93 & 85.97 & 75.60 & 81.13 & 165.75 & 1.95 \\
    E12 & 4o-mini FT-M & --- & Minimal fine-tune & 86.55 & 87.00 & 89.70 & 89.58 & \cellcolor{lightgray}86.97 & \cellcolor{lightgray}87.08 & 75.62 & 82.76 & \cellcolor{lightgray}16.60 & \cellcolor{lightgray}0.19 \\
    E13 & 4o-mini FT & Base FT & Prompt, Label, FT w/prompt & 81.42 & 81.90 & 84.77 & 84.78 & 74.49 & 74.72 & 70.95 & 79.55 & 42.88 & 0.53 \\
    E14 & 4o-mini FT-L & Base FT & Prompt, Label, FT w/prompt, label & 82.02 & 82.53 & 85.24 & 85.11 & 78.15 & 78.47 & 71.68 & 79.64 & 49.31 & 0.60 \\
    E15 & 4o-mini FT-L 5 & Base FT & Prompt, Label, FT w/prompt, label (5 epochs) & 83.88 & 84.27 & 86.74 & 86.61 & 81.62 & 81.81 & 75.00 & 81.27 & 207.64 & 2.48 \\
    E16 & 4o-mini FT & Large FT & Fine-tune w/prompt & 84.00 & 84.58 & 87.65 & 87.58 & 80.37 & 80.69 & 72.46 & 80.95 & 86.41 & 1.03 \\
    E17 & 4o-mini FT-L & Large FT & Fine-tune w/prompt, label & 84.16 & 84.70 & 87.65 & 87.56 & 80.89 & 81.11 & 73.29 & 81.22 & 92.84 & 1.10 \\
    E18 & 4o-mini FT-L 5 & Large FT & Fine-tune w/prompt, label (5 epochs) & 84.87 & 85.25 & 87.87 & 87.75 & 83.80 & 83.89 & 75.58 & 81.63 & 251.17 & 2.96 \\
    \midrule
      E19 & 4o & Large FT & Prompt, Label & \cellcolor{lightgray}83.18 & 83.68 & 85.71 & 85.56 & \cellcolor{lightgray}81.98 & \cellcolor{lightgray}82.22 & \cellcolor{lightgray}73.44 & \cellcolor{lightgray}81.09 & \cellcolor{lightgray}\textbf{53.26} & \cellcolor{lightgray}\textbf{0.64} \\
    E20 & 4o & Large FT & Prompt, Label, Examples (Few shot) & 83.09 & 83.66 & 86.01 & 85.86 & 81.53 & 81.81 & 72.06 & 80.68 & \cellcolor{lightgray}53.26 & \cellcolor{lightgray}0.64 \\
    E21 & 4o & Large FT & Prompt, Label, Balanced Ex. (Few shot) & 82.99 & 83.55 & 85.87 & 85.69 & 80.89 & 81.11 & 72.15 & 80.86 & \cellcolor{lightgray}53.26 & \cellcolor{lightgray}0.64 \\
    E22 & 4o & Large FT & Prompt, Probs & 82.65 & 83.25 & 86.05 & 85.97 & 77.67 & 77.92 & 71.16 & 80.54 & \cellcolor{lightgray}53.26 & \cellcolor{lightgray}0.64 \\
    E23 & 4o & Large FT & Prompt, Label, Probs & 83.08 & \cellcolor{lightgray}83.71 & \cellcolor{lightgray}86.44 & \cellcolor{lightgray}86.33 & 79.23 & 79.58 & 71.48 & 80.77 & \cellcolor{lightgray}53.26 & \cellcolor{lightgray}0.64 \\
    E24 & 4o & Large FT & Prompt, Label, Probs, Examples & 82.74 & 83.35 & 86.32 & 86.22 & 77.76 & 78.06 & 70.98 & 80.41 & \cellcolor{lightgray}53.26 & \cellcolor{lightgray}0.64 \\
    \midrule
      E25 & 4o FT & --- & Fine-tune w/prompt & 86.83 & 87.43 & 90.44 & 90.36 & 88.34 & 88.47 & 73.08 & \cellcolor{lightgray}82.31 & 276.24 & 3.18 \\
    E26 & 4o FT-M & --- & Minimal fine-tune & \cellcolor{lightgray}\textbf{86.99} & \cellcolor{lightgray}\textbf{87.57} & \cellcolor{lightgray}\textbf{90.57} & \cellcolor{lightgray}\textbf{90.50} & \cellcolor{lightgray}\textbf{89.00} & \cellcolor{lightgray}\textbf{89.17} & \cellcolor{lightgray}73.99 & 82.26 & \cellcolor{lightgray}138.37 & \cellcolor{lightgray}1.59 \\
    E27 & 4o FT & Large FT & Fine-tune w/prompt & 83.82 & 84.47 & 87.80 & 87.72 & 79.49 & 79.86 & 71.18 & 80.68 & 329.50 & 3.93 \\
    E28 & 4o FT-L & Large FT & Fine-tune w/prompt, label & 84.23 & 84.82 & 87.74 & 87.64 & 80.55 & 80.83 & 72.63 & 81.54 & 383.10 & 4.55 \\
    \bottomrule
    \end{tabular}
    
    \label{tab:experiment_summary_round1}
    \end{threeparttable}
\end{table}

\begin{table}[!ht]
    \begin{threeparttable}
    \centering
    \setlength{\tabcolsep}{3pt}
    \tiny
    \caption{Round 2 Results}
    \begin{tabular}{l l l l r r r r r r r r r r}
    \toprule
    & & & & \multicolumn{2}{c}{Merged} & \multicolumn{2}{c}{DynaSent R1} & \multicolumn{2}{c}{DynaSent R2} & \multicolumn{2}{c}{SST-3} & \multicolumn{2}{c}{Cost (\$)} \\
    \cmidrule(lr){5-6} \cmidrule(lr){7-8} \cmidrule(lr){9-10} \cmidrule(lr){11-12} \cmidrule(lr){13-14}
    ID & GPT\tnote{1} & ELECTRA & Description & F1 & Acc & F1 & Acc & F1 & Acc & F1 & Acc & FT & /F1 \\
    \midrule
      B1 & --- & Base & Baseline, Classifier head & 69.37 & 69.57 & 70.75 & 71.11 & 61.19 & 61.25 & 60.34 & 69.77 & 0.65 & 0.01 \\
    B2 & --- & Large & Baseline, Classifier head & 67.99 & 68.15 & 69.67 & 70.03 & 59.78 & 59.72 & 58.21 & 67.83 & 2.51 & 0.04 \\
    B3 & 4o-mini & --- & Baseline (Zero shot) & 79.29 & 80.15 & \cellcolor{lightgray}81.19 & \cellcolor{lightgray}81.08 & 76.69 & 77.36 & 69.30 & 79.55 & --- & --- \\
    B4 & 4o & --- & Baseline (Zero shot) & \cellcolor{lightgray}79.80 & \cellcolor{lightgray}80.47 & 80.77 & 80.61 & \cellcolor{lightgray}80.05 & \cellcolor{lightgray}80.42 & \cellcolor{lightgray}71.96 & \cellcolor{lightgray}80.27 & --- & --- \\
    \midrule
      E1 & --- & Base FT & Fine-tune all layers & 78.98 & 79.46 & 82.13 & 82.19 & 69.51 & 69.72 & 68.13 & 78.19 & \cellcolor{lightgray}9.73 & \cellcolor{lightgray}0.12 \\
    E2 & --- & Large FT & Fine-tune all layers & \cellcolor{lightgray}83.16 & \cellcolor{lightgray}83.71 & \cellcolor{lightgray}86.53 & \cellcolor{lightgray}86.44 & \cellcolor{lightgray}78.36 & \cellcolor{lightgray}78.61 & \cellcolor{lightgray}72.63 & \cellcolor{lightgray}80.91 & 53.26 & 0.64 \\
    \midrule
      E3 & 4o-mini & Base FT & Prompt, Label & 82.26 & 82.92 & 86.29 & 86.25 & 74.47 & 74.86 & 70.03 & 80.14 & \cellcolor{lightgray}9.73 & \cellcolor{lightgray}0.12 \\
    E4 & 4o-mini & Large FT & Prompt, Label & \cellcolor{lightgray}84.10 & \cellcolor{lightgray}84.73 & \cellcolor{lightgray}87.90 & \cellcolor{lightgray}87.83 & 79.52 & 79.86 & 72.54 & \cellcolor{lightgray}81.27 & 53.26 & 0.63 \\
    E5 & 4o-mini & Large FT & Prompt, Label, Examples (Few shot) & 83.63 & 84.18 & 87.13 & 87.03 & \cellcolor{lightgray}80.29 & \cellcolor{lightgray}80.56 & 72.68 & 80.72 & 53.26 & 0.64 \\
    E6 & 4o-mini & Large FT & Prompt, Label, Balanced Ex. (Few shot) & 83.28 & 83.81 & 86.71 & 86.58 & 80.12 & 80.42 & 72.51 & 80.41 & 53.26 & 0.64 \\
    E7 & 4o-mini & Large FT & Prompt, Probs & 83.53 & 84.07 & 86.76 & 86.67 & 79.89 & 80.14 & \cellcolor{lightgray}72.98 & 81.13 & 53.26 & 0.64 \\
    E8 & 4o-mini & Large FT & Prompt, Label, Probs & 83.88 & 84.47 & 87.41 & 87.33 & 80.23 & \cellcolor{lightgray}80.56 & 72.53 & 81.09 & 53.26 & 0.64 \\
    E9 & 4o-mini & Large FT & Prompt, Label, Probs, Examples & 83.46 & 84.03 & 87.00 & 86.89 & 79.28 & 79.58 & 72.39 & 80.81 & 53.26 & 0.64 \\
    \midrule
      E10 & 4o-mini FT & --- & Fine-tune w/prompt & \cellcolor{lightgray}86.62 & \cellcolor{lightgray}87.09 & \cellcolor{lightgray}89.44 & \cellcolor{lightgray}89.33 & 87.08 & 87.22 & \cellcolor{lightgray}\textbf{75.98} & \cellcolor{lightgray}\textbf{83.39} & 33.15 & 0.38 \\
    E11 & 4o-mini FT 5 & --- & Fine-tune w/prompt (5 epochs) & 84.76 & 85.07 & 87.64 & 87.47 & 86.50 & 86.53 & 75.15 & 80.68 & 165.75 & 1.96 \\
    E12 & 4o-mini FT-M & --- & Minimal fine-tune & 86.47 & 86.89 & \cellcolor{lightgray}89.44 & 89.31 & \cellcolor{lightgray}87.28 & \cellcolor{lightgray}87.36 & 75.86 & 82.81 & \cellcolor{lightgray}16.60 & \cellcolor{lightgray}0.19 \\
    E13 & 4o-mini FT & Base FT & Prompt, Label, FT w/prompt & 80.69 & 81.24 & 84.57 & 84.58 & 71.62 & 71.94 & 68.45 & 78.82 & 42.88 & 0.53 \\
    E14 & 4o-mini FT-L & Base FT & Prompt, Label, FT w/prompt, label & 81.65 & 82.25 & 85.15 & 85.03 & 76.43 & 76.81 & 69.71 & 79.50 & 49.31 & 0.60 \\
    E15 & 4o-mini FT-L 5 & Base FT & Prompt, Label, FT w/prompt, label (5 epochs) & 83.46 & 83.89 & 86.02 & 85.89 & 80.75 & 80.97 & 75.04 & 81.58 & 207.64 & 2.49 \\
    E16 & 4o-mini FT & Large FT & Fine-tune w/prompt & 83.87 & 84.44 & 87.49 & 87.42 & 79.97 & 80.28 & 72.46 & 80.95 & 86.41 & 1.03 \\
    E17 & 4o-mini FT-L & Large FT & Fine-tune w/prompt, label & 84.07 & 84.59 & 87.51 & 87.42 & 80.61 & 80.83 & 73.38 & 81.22 & 92.84 & 1.10 \\
    E18 & 4o-mini FT-L 5 & Large FT & Fine-tune w/prompt, label (5 epochs) & 84.79 & 85.15 & 87.63 & 87.50 & 84.94 & 85.00 & 75.59 & 81.36 & 251.17 & 2.96 \\
    \midrule
      E19 & 4o & Large FT & Prompt, Label & 83.20 & 83.69 & 85.71 & 85.56 & \cellcolor{lightgray}82.13 & \cellcolor{lightgray}82.36 & 73.52 & 81.09 & \cellcolor{lightgray}53.26 & \cellcolor{lightgray}0.64 \\
    E20 & 4o & Large FT & Prompt, Label, Examples (Few shot) & 83.48 & 84.00 & 86.21 & 86.06 & 81.43 & 81.67 & \cellcolor{lightgray}73.85 & \cellcolor{lightgray}81.40 & \cellcolor{lightgray}53.26 & \cellcolor{lightgray}0.64 \\
    E21 & 4o & Large FT & Prompt, Label, Balanced Ex. (Few shot) & 83.38 & 83.89 & 86.14 & 85.97 & 81.18 & 81.39 & 73.61 & 81.31 & \cellcolor{lightgray}53.26 & \cellcolor{lightgray}0.64 \\
    E22 & 4o & Large FT & Prompt, Probs & 83.32 & 83.87 & 86.68 & 86.58 & 79.16 & 79.44 & 72.63 & 80.90 & \cellcolor{lightgray}53.26 & \cellcolor{lightgray}0.64 \\
    E23 & 4o & Large FT & Prompt, Label, Probs & \cellcolor{lightgray}83.54 & \cellcolor{lightgray}84.10 & \cellcolor{lightgray}86.93 & \cellcolor{lightgray}86.81 & 79.69 & 80.00 & 72.85 & 81.04 & \cellcolor{lightgray}53.26 & \cellcolor{lightgray}0.64 \\
    E24 & 4o & Large FT & Prompt, Label, Probs, Examples & 83.34 & 83.89 & 86.73 & 86.61 & 79.18 & 79.44 & 72.68 & 80.90 & \cellcolor{lightgray}53.26 & \cellcolor{lightgray}0.64 \\
    \midrule
      E25 & 4o FT & --- & Fine-tune w/prompt & 86.74 & 87.32 & 90.48 & 90.42 & 87.94 & 88.06 & 73.09 & 82.04 & 276.24 & 3.18 \\
    E26 & 4o FT-M & --- & Minimal fine-tune & \cellcolor{lightgray}\textbf{86.99} & \cellcolor{lightgray}\textbf{87.57} & \cellcolor{lightgray}\textbf{90.57} & \cellcolor{lightgray}\textbf{90.50} & \cellcolor{lightgray}\textbf{89.00} & \cellcolor{lightgray}\textbf{89.17} & \cellcolor{lightgray}73.99 & \cellcolor{lightgray}82.26 & \cellcolor{lightgray}138.37 & \cellcolor{lightgray}1.59 \\
    E27 & 4o FT & Large FT & Fine-tune w/prompt & 84.24 & 84.84 & 87.99 & 87.89 & 80.52 & 80.83 & 72.81 & 81.18 & 329.50 & 3.91 \\
    E28 & 4o FT-L & Large FT & Fine-tune w/prompt, label & 84.50 & 85.04 & 87.87 & 87.75 & 82.01 & 82.22 & 73.56 & 81.54 & 383.10 & 4.53 \\
    \bottomrule
    \end{tabular}
    
    \begin{tablenotes}[flushleft]
    \tiny
    \item \textbf{Bold} = best overall, \colorbox{lightgray}{highlighted} = best in section 
    \item\textsuperscript{1} GPT fine-tuning types: FT = fine-tune all layers with prompt, FT-M = minimal fine-tune format without prompt, FT-L = fine-tune with prompt including ELECTRA label, FT 5 = fine-tune for 5 epochs
    \end{tablenotes}
    
    \label{tab:experiment_summary_round2}
    \end{threeparttable}
\end{table}

\clearpage
\section{ELECTRA Fine-tuning Details}
\label{app:finetune}
\begin{table}[h]

    \centering
    \setlength{\tabcolsep}{8pt}
    \small
    \caption{ELECTRA Fine-Tune Configuration}
    \begin{tabular}{l r r}
    \toprule
    Setting & ELECTRA Base FT & ELECTRA Large FT \\
    \midrule
    Source & Hugging Face & Hugging Face \\
    Source Model ID & google/electra-base-discriminator & google/electra-large-discriminator \\
    Encoder Blocks & 12 & 24 \\
    Embedding Dimension & 768 & 1024 \\
    Attention Heads & 12 & 16 \\
    Feedforward Size & 3072 & 4096 \\
    Parameters & 110 Million & 335 Million \\
    \midrule
    Custom Pooling Layer Method & Mean & Mean \\
    Classifier Head Hidden Layers & 2 & 2 \\
    Classifier Head Hidden Dimension & 1024 & 1024 \\
    Classifier Head Hidden Activation & SwishGLU & SwishGLU \\
    \midrule
    Finetuned Encoder Blocks & 12 & 24 \\
    Total Layers & 104 & 200 \\
    Total Parameters & 112,830,979 & 338,293,763 \\
    Trainable Parameters & 100\% & 100\% \\
    \midrule
    Learning Rate & $1e^{-5}$ & $1e^{-5}$ \\
    Learning Rate Decay & 0.95 & 0.95 \\
    Batch Size & 16 & 16 \\
    Accumulation Steps & 2 & 2 \\
    Target Epochs & 50 & 50 \\
    Actual Epochs & 20 & 23 \\
    Selected Best Epoch & 14 & 13 \\
    Dropout Rate & 0.30 & 0.30 \\
    L2 Strength & 0.01 & 0.01 \\
    Optimizer & AdamW & AdamW \\
    Zero Redundancy & Yes & Yes \\
    Scheduler & CosineAnnealingWarmRestarts& CosineAnnealingWarmRestarts \\
    Scheduler: T\_0 & 5 & 5 \\
    Scheduler: T\_mult & 1 & 1 \\
    Scheduler: eta\_min & $1e^{-7}$ & $1e^{-7}$ \\
    Early Stop & Validation F1 Score & Validation F1 Score \\
    N Iterations No Change & 10 & 10 \\
    \midrule
    Dataset & Merged (Dyn R1, Dyn R2, SST-3) & Merged (Dyn R1, Dyn R2, SST-3) \\
    Train Size & 102,097 & 102,097 \\
    Train Label Distribution & Neu: 49,148, Pos: 31,039, Neg: 21,910 & Neu: 49,148, Pos: 31,039, Neg: 21,910 \\
    Validation Size & 5,421 & 5,421 \\
    Validation Label Distribution & Neu: 1,669, Pos: 1,884, Neg: 1,868 & Neu: 1,669, Pos: 1,884, Neg: 1,868 \\
    \midrule
    Hosting Provider & Lambda Labs & Lambda Labs \\
    GPU Type & Tesla V100 & A100 \\
    GPU Memory & 16 GB & 40 GB \\
    GPU Quantity & 8 & 8 \\
    Rate & \$4.40/hour & \$10.32/hour \\
    Training Time (Up to Selected Epoch) & 02:12:44 & 05:09:23 \\
    Training Time (Total) & 03:09:40 & 09:23:29 \\
    Cost (Up to Selected Epoch) & \$9.73 & \$53.26 \\
    Cost (Total) & \$13.91 & \$96.92 \\
    \bottomrule
    \end{tabular}
    \label{tab:electra_finetune_configuration}
\end{table}

\clearpage
\begin{landscape}
\section{GPT Fine-tuning Details}
\label{app:finetune2}
\begin{table}[h]
    \centering
    \setlength{\tabcolsep}{6pt} 
    \small
    \caption{GPT Fine-Tune Configuration}
    \begin{tabular}{l p{2.0cm} p{2.0cm} p{2.0cm} p{2.0cm} p{2.0cm} p{2.0cm} p{2.0cm} p{2.0cm} p{2.0cm}}
    \toprule
    Setting & 4o-mini\newline FT-M & 4o-mini\newline FT & 4o-mini\newline FT-L Base 1 & 4o-mini\newline FT-L Base 5 & 4o-mini\newline FT-L Large 1 & 4o-mini\newline FT-L Large 5 & 4o\newline FT-M & 4o\newline FT & 4o\newline FT-L Large 1 \\
    \midrule
    GPT Model & 4o-mini & 4o-mini & 4o-mini & 4o-mini & 4o-mini & 4o-mini & 4o & 4o & 4o \\
    ELECTRA Model & None & None & Base FT & Base FT & Base FT & Base FT & None & None & Large FT \\
    Code & FT-M & FT & FT-L & FT-L & FT-L & FT-L & FT-M & FT & FT-L \\
    Code Meaning & Minimal\newline Format & Fine-Tune & Fine-Tune w/Label & Fine-Tune w/Label & Fine-Tune w/Label & Fine-Tune w/Label & Minimal\newline Format & Fine-Tune & Fine-Tune w/Label \\
    Format & JSON\newline (No Prompt) & DSPy Prompt & DSPy Prompt + Label & DSPy Prompt + Label & DSPy Prompt + Label & DSPy Prompt + Label & JSON\newline (No Prompt) & DSPy Prompt & DSPy Prompt + Label \\
    \midrule
    Source & OpenAI & OpenAI & OpenAI & OpenAI & OpenAI & OpenAI & OpenAI & OpenAI & OpenAI \\
    Source Model ID & gpt-4o-mini-2024-07-18 & gpt-4o-mini-2024-07-18 & gpt-4o-mini-2024-07-18 & gpt-4o-mini-2024-07-18 & gpt-4o-mini-2024-07-18 & gpt-4o-mini-2024-07-18 & gpt-4o-2024-08-06 & gpt-4o-2024-08-06 & gpt-4o-2024-08-06 \\
    \midrule
    Dataset & Merged & Merged & Merged & Merged & Merged & Merged & Merged & Merged & Merged \\
    Train Size & 102,097 & 102,097 & 102,097 & 102,097 & 102,097 & 102,097 & 102,097 & 102,097 & 102,097 \\
    Validation Size & 5,421 & 5,421 & 5,421 & 5,421 & 5,421 & 5,421 & 5,421 & 5,421 & 5,421 \\
    \midrule
    Fine-Tuning Date & 2024-10-23 &  &  &  &  &  &  &  &  \\
    Total Job Time & 01:17:07 & 01:07:05 & 01:09:00 & 01:09:00 & 01:09:00 & 01:09:00 & 01:31:06 & 01:43:43 & 01:47:49 \\
    \midrule
    LR Multiplier & 1.8 & 1.8 & 1.8 & 1.8 & 1.8 & 1.8 & 2.0 & 2.0 & 2.0 \\
    Seed & 42 & 42 & 42 & 42 & 42 & 42 & 42 & 42 & 42 \\
    Batch Size & 68 & 68 & 68 & 68 & 68 & 68 & 68 & 68 & 68 \\
    Epochs & 1 & 1 & 1 & 1 & 1 & 1 & 1 & 1 & 1 \\
    \midrule
    Train Loss & 0.1250 & 0.1383 & 0.1286 & 0.1286 & 0.1286 & 0.1286 & 0.1185 & 0.1150 & 0.1075 \\
    Validation Loss & 0.1112 & 0.1114 & 0.1429 & 0.1429 & 0.1429 & 0.1429 & 0.1054 & 0.1056 & 0.1242 \\
    \midrule
    Trained Tokens & 5,534,849 & 11,049,720 & 13,193,757 & 13,193,757 & 13,193,757 & 13,193,757 & 5,534,849 & 11,049,720 & 13,193,757 \\
    Rate & \$3/1M tokens & \$3/1M tokens & \$3/1M tokens & \$3/1M tokens & \$3/1M tokens & \$3/1M tokens & \$25/1M tokens & \$25/1M tokens & \$25/1M tokens \\
    Cost & \$16.60 & \$33.15 & \$39.58 & \$39.58 & \$39.58 & \$39.58 & \$138.37 & \$276.24 & \$329.84 \\
    \bottomrule
    \end{tabular}
    \label{tab:gpt_finetune_configuration}
\end{table}
\end{landscape}

\clearpage\section{OpenAI Fine-tuning Templates}
\label{app:ft_templates}

\begin{table*}[ht!]
    \centering
    \setlength{\tabcolsep}{12pt}
    \small
    \caption{FT-M: Minimal Template for Fine-tuning}
    \begin{tabular}{l p{13cm}}
    \toprule
    Role & Content \\
    \midrule
    System & ``You are a model that classifies the sentiment of a review as either `positive', `neutral', or `negative'.'' \\
    User & ``Those 2 drinks are part of the HK culture and has years of history. It is so bad.'' \\
    Assistant & ``negative'' \\
    \bottomrule
    \end{tabular}
    \label{tab:jsonl_minimal}
\end{table*}

\begin{table*}[ht!]
    \centering
    \setlength{\tabcolsep}{12pt}
    \small
    \caption{FT: Prompt Template for Fine-tuning}
    \begin{tabular}{l p{13cm}}
    \toprule
    Role & Content \\
    \midrule
    System & ``You are a sentiment analysis assistant.'' \\
    User & ``\textbf{Classify the sentiment of a review as either `negative', `neutral', or `positive'.\textbackslash n---\textbackslash n\textbackslash nFollow the following format.\textbackslash n\textbackslash nReview: The review text to classify.\textbackslash nClassification: One word representing the sentiment classification: `negative', `neutral', or `positive' (do not repeat the field name, do not use `mixed')\textbackslash n\textbackslash n---\textbackslash n\textbackslash nReview:} Those 2 drinks are part of the HK culture and has years of history. It is so bad.\textbackslash n\textbf{Classification:}'' \\
    Assistant & ``negative'' \\
    \bottomrule
    \end{tabular}
    \label{tab:jsonl_prompt}
\end{table*}

\begin{table*}[ht!]
    \centering
    \setlength{\tabcolsep}{12pt}
    \small
    \caption{FT-L: Prompt with Predicted Label Template for Fine-tuning}
    \begin{tabular}{l p{13cm}}
    \toprule
    Role & Content \\
    \midrule
    System & ``You are a sentiment analysis assistant.'' \\
    User & ``Classify the sentiment of a review as either `negative', `neutral', or `positive'.\textbackslash n---\textbackslash n\textbackslash nFollow the following format.\textbackslash n\textbackslash nReview: The review text to classify.\textbackslash n\textbf{Classifier Decision: The sentiment classification proposed by a model fine-tuned on sentiment.}\textbackslash nClassification: One word representing the sentiment classification: `negative', `neutral', or `positive' (do not repeat the field name, do not use `mixed')\textbackslash n\textbackslash n---\textbackslash n\textbackslash nReview: Those 2 drinks are part of the HK culture and has years of history. It is so bad.\textbackslash n\textbf{Classifier Decision: negative}\textbackslash nClassification:'' \\
    Assistant & ``negative'' \\
    \bottomrule
    \end{tabular}
    \label{tab:jsonl_promptpred}
\end{table*}

\clearpage\section{DSPy Prompt Signature Examples}
\label{app:prompts}

\begin{figure}[h]
    \centering
    \caption{Basic Prompt DSPy Signature}
    \label{fig:prompt-basic}
    \scriptsize
    \begin{tcolorbox}[
        colback=white,
        colframe=black,
        boxrule=0.5pt,
        arc=0pt,
        outer arc=0pt,
        left=5pt,
        right=5pt,
        top=5pt,
        bottom=5pt
    ]
\begin{verbatim}
Classify the sentiment of a review as either 'negative', 'neutral', or 'positive'.

---

Follow the following format.

Review: The review text to classify.
Classification: One word representing the sentiment classification: 'negative', 'neutral', or 'positive'
(do not repeat the field name, do not use 'mixed').

---

Review: Those 2 drinks are part of the HK culture and has years of history. It is so bad.
Classification:
\end{verbatim}
\end{tcolorbox}
\end{figure}

\begin{figure}[h]
    \centering
    \caption{Prompt with Predicted Label DSPy Signature}
    \label{fig:prompt-label}
    \scriptsize
    \begin{tcolorbox}[
        colback=white,
        colframe=black,
        boxrule=0.5pt,
        arc=0pt,
        outer arc=0pt,
        left=5pt,
        right=5pt,
        top=5pt,
        bottom=5pt
    ]
\begin{verbatim}
Classify the sentiment of a review as either 'negative', 'neutral', or 'positive'.

---

Follow the following format.

Review: The review text to classify.
Classifier Decision: The sentiment classification proposed by a model fine-tuned on sentiment.
Classification: One word representing the sentiment classification: 'negative', 'neutral', or 'positive'
(do not repeat the field name, do not use 'mixed')

---

Review: I was told by the repair company that was doing the car repair that fixing the rim was
"impossible" and to replace it.
Classifier Decision: negative
Classification:
\end{verbatim}
\end{tcolorbox}
\end{figure}

\begin{figure}[h]
    \centering
    \caption{Prompt with Probabilities DSPy Signature}
    \label{fig:prompt-probs}
    \scriptsize
    \begin{tcolorbox}[
        colback=white,
        colframe=black,
        boxrule=0.5pt,
        arc=0pt,
        outer arc=0pt,
        left=5pt,
        right=5pt,
        top=5pt,
        bottom=5pt
    ]
\begin{verbatim}
Classify the sentiment of a review as either 'negative', 'neutral', or 'positive'.

---

Follow the following format.

Review: The review text to classify.

Negative Probability: Probability the review is negative from a model fine-tuned on sentiment

Neutral Probability: Probability the review is neutral from a model fine-tuned on sentiment

Positive Probability: Probability the review is positive from a model fine-tuned on sentiment

Classification: One word representing the sentiment classification: 'negative', 'neutral', or 'positive'
(do not repeat the field name, do not use 'mixed')

---

Review: Those 2 drinks are part of the HK culture and has years of history. It is so bad.

Negative Probability: 99.85%

Neutral Probability: 0.04%

Positive Probability: 0.12%

Classification:
\end{verbatim}
\end{tcolorbox}
\end{figure}

\begin{figure}[h]
    \centering
    \caption{Prompt with Predicted Label and Probabilities DSPy Signature}   
    \label{fig:prompt-labelprobs}
    \scriptsize
    \begin{tcolorbox}[
        colback=white,
        colframe=black,
        boxrule=0.5pt,
        arc=0pt,
        outer arc=0pt,
        left=5pt,
        right=5pt,
        top=5pt,
        bottom=5pt
    ]
\begin{verbatim}
Classify the sentiment of a review as either 'negative', 'neutral', or 'positive'.

---

Follow the following format.

Review: The review text to classify.

Classifier Decision: The sentiment classification proposed by a model fine-tuned on sentiment.

Negative Probability: Probability the review is negative

Neutral Probability: Probability the review is neutral

Positive Probability: Probability the review is positive

Classification: One word representing the sentiment classification: 'negative', 'neutral', or 'positive'
(do not repeat the field name, do not use 'mixed')

---

Review: Those 2 drinks are part of the HK culture and has years of history. It is so bad.

Classifier Decision: negative

Negative Probability: 99.85%

Neutral Probability: 0.04%

Positive Probability: 0.12%

Classification:
\end{verbatim}
\end{tcolorbox}
\end{figure}

\begin{figure}[h]
    \centering
    \caption{Top Examples DSPy Signature}
    \label{fig:prompt-examples}
    \scriptsize
    \begin{tcolorbox}[
        colback=white,
        colframe=black,
        boxrule=0.5pt,
        arc=0pt,
        outer arc=0pt,
        left=5pt,
        right=5pt,
        top=5pt,
        bottom=5pt
    ]
\begin{verbatim}
Classify the sentiment of a review as either 'negative', 'neutral', or 'positive'.

---

Follow the following format.

Examples: A list of examples that demonstrate different sentiment classes.

Review: The review text to classify.

Classifier Decision: The sentiment classification proposed by a model fine-tuned on sentiment.

Classification: One word representing the sentiment classification: 'negative', 'neutral', or 'positive'
(do not repeat the field name, do not use 'mixed')

---

Examples:
- negative: We've been to about 5 or 6 other Verizon stores in Vegas, and they all give us a hard time
about everything and never solve any issue.
- negative: Then Raj then had the balls to send me an email after my box was closed to tell me they were
ready to receive the key for my mailbox after closing it.!
- negative: Always and issue here even with take out orders.
- negative: SHOULD YOU HAVE ANY DISPUTE, THEY IMMEDIATELY WILL THREATEN YOU WITH MECHANICS LIENS.
- negative: We were waiting for them to get our order out, but the lady came out and gave the car behind
us their order first!

Review: I went back in to ask for cilantro dressing the shift leader even smile or greet me.

Classifier Decision: negative

Classification:
\end{verbatim}
\end{tcolorbox}
\end{figure}

\begin{figure}[hb]
    \centering
    \caption{Balanced Examples DSPy Signature}
    \label{fig:prompt-balanced}
    \scriptsize
    \begin{tcolorbox}[
        colback=white,
        colframe=black,
        boxrule=0.5pt,
        arc=0pt,
        outer arc=0pt,
        left=5pt,
        right=5pt,
        top=5pt,
        bottom=5pt
    ]
\begin{verbatim}
Classify the sentiment of a review as either 'negative', 'neutral', or 'positive'.

---

Follow the following format.

Examples: A list of examples that demonstrate different sentiment classes.

Review: The review text to classify.

Classifier Decision: The sentiment classification proposed by a model fine-tuned on sentiment.

Classification: One word representing the sentiment classification: 'negative', 'neutral', or 'positive'
(do not repeat the field name, do not use 'mixed')

---

Examples:
- negative: Beware of all the fake 5 star reviews of this place, just take a look at these people.
- negative: 3- girls look even cheaper than the club.
- neutral: Not to mention the esso across the street also has cheaper gas.
- neutral: I wish that they would open up by 6am so that I can pick up a coffee or tea before work, but
what boba place is opened that early?
- positive: The plumbers did not give up and continued to work on the drain for two days.
- positive: This is my 6th gun to add to my collection and if I had not wanted it so bad, I would have
walked out 2 minutes after walking in.

Review: She greeted customers by holding the scanner toward them without even looking.

Classifier Decision: negative

Classification:
\end{verbatim}
\end{tcolorbox}
\end{figure}

\begin{figure}[h]
    \centering
    \caption{All Context DSPy Signature}
    \label{fig:prompt-all}
    \scriptsize
    \begin{tcolorbox}[
        colback=white,
        colframe=black,
        boxrule=0.5pt,
        arc=0pt,
        outer arc=0pt,
        left=5pt,
        right=5pt,
        top=5pt,
        bottom=5pt
    ]
\begin{verbatim}
Classify the sentiment of a review as either 'negative', 'neutral', or 'positive'.

---

Follow the following format.

Examples: A list of examples that demonstrate different sentiment classes.

Review: The review text to classify.

Classifier Decision: The sentiment classification proposed by a model fine-tuned on sentiment.

Negative Probability: Probability the review is negative

Neutral Probability: Probability the review is neutral

Positive Probability: Probability the review is positive

Classification: One word representing the sentiment classification: 'negative', 'neutral', or 'positive'
(do not repeat the field name, do not use 'mixed')

---

Examples:
- negative: The only negative I can think for this place is it's price-point.
- positive: This place will be the death of my waist (but not my wallet).
- negative: Expensive, if you are looking for something more affordable, don't go here; you will miss
  the best dishes.
- positive: Thank you so much for dealing with my crabby ass
- positive: I think I scarfed it down so quickly because it was that good! It was bad.

Review: The gentleman staffing the bar seemed a bit gruff, but a good caffeine fix will help me forgive
even the orneriest grump.

Classifier Decision: negative

Negative Probability: 84.37%

Neutral Probability: 0.53%

Positive Probability: 15.10%

Classification:
\end{verbatim}
\end{tcolorbox}
\end{figure}

\end{document}